\documentclass[conference,a4paper,10pt]{IEEEtran}
\IEEEoverridecommandlockouts
\usepackage{cite}
\usepackage{amsmath,amssymb,amsfonts}
\usepackage{graphicx}
\usepackage{booktabs}
\usepackage{array}
\usepackage{multirow}
\usepackage{url}
\usepackage{textcomp}
\usepackage{xcolor}
\usepackage{algorithm}
\usepackage{algpseudocode}
\usepackage[hidelinks]{hyperref}

\algrenewcommand\algorithmicindent{0.9em}

\begin{document}

\title{AB-RAG: Adaptive Budgeted Retrieval-Augmented Generation\\
for Reliable Question Answering}

\author{\IEEEauthorblockN{Ansh Kamthan}
\IEEEauthorblockA{\textit{Department of Artificial Intelligence and Machine Learning}\\
\textit{Manipal University Jaipur}\\
Jaipur, India\\
anshkamthan4@gmail.com}}

\maketitle
\thispagestyle{plain}
\pagestyle{plain}

\begin{abstract}
Retrieval-Augmented Generation (RAG) has become the standard way to ground large
language models in external knowledge, yet most systems retrieve a fixed number of
passages for every question regardless of its difficulty. This wastes computation on
easy questions, starves hard ones, and gives no signal for when a generated answer
can be trusted. With a growing share of question answering systems built on top of
commercial language model APIs, a method that can decide how much to retrieve, and
how far to trust its own answers, without retraining the underlying model, is of
clear practical value. This paper presents AB-RAG (Adaptive Budgeted
Retrieval-Augmented Generation), a training-free and backbone-agnostic framework that
generates an answer, estimates its confidence from a combination of three signals,
and then decides whether to stop or to retrieve more evidence, subject to a fixed
retrieval budget. The estimator combines the model's own certainty, the agreement
between the answer and the evidence, and the variance of the retrieval scores. For
models that expose token probabilities the certainty signal is read directly; for
closed APIs it is approximated by self-consistency, so the method works without access
to model internals. Across three backbones and two datasets, the central result is
that the confidence estimate reliably separates correct from incorrect answers on
every backbone, reaching a clean split of 57.6\% against 0\% Exact Match between high-
and low-confidence answers on a factoid dataset. The adaptive policy improves accuracy
on capable backbones, and the study reports its negative and nuanced findings honestly,
including a confidence signal that proved unsuitable for short answers and a retrieval
signal whose sign was found and corrected through measurement. The entire study was
carried out on a single consumer laptop with only a few dollars of API spend.
\end{abstract}


\section{Introduction}
Retrieval-Augmented Generation (RAG)~\cite{lewis2020rag} has become the standard way to
ground large language models (LLMs) in external knowledge. Instead of relying only on
the facts a model memorised during training, a RAG system retrieves relevant passages
from a corpus and gives them to the model as evidence before it answers. A typical
pipeline retrieves a fixed number of passages for every query, attaches them to the
prompt, and generates an answer. This works well in many cases, but it carries a hidden
assumption that every question needs the same amount of evidence, and that assumption is
usually wrong. A simple factoid question such as ``Of which African country is Niamey
the capital?'' can be answered from a single passage. A multi-hop question such as
``Which director links Film A and Actor B?'' often needs several rounds of evidence
gathering to connect the intermediate facts. When the retrieval depth is fixed, the
system over-retrieves for easy questions, wasting computation, context-window space,
and, on paid APIs, money. At the same time it under-retrieves for hard questions and
leaves them without enough evidence to answer correctly.

\begin{figure*}[t]
\centering
\includegraphics[width=0.96\textwidth]{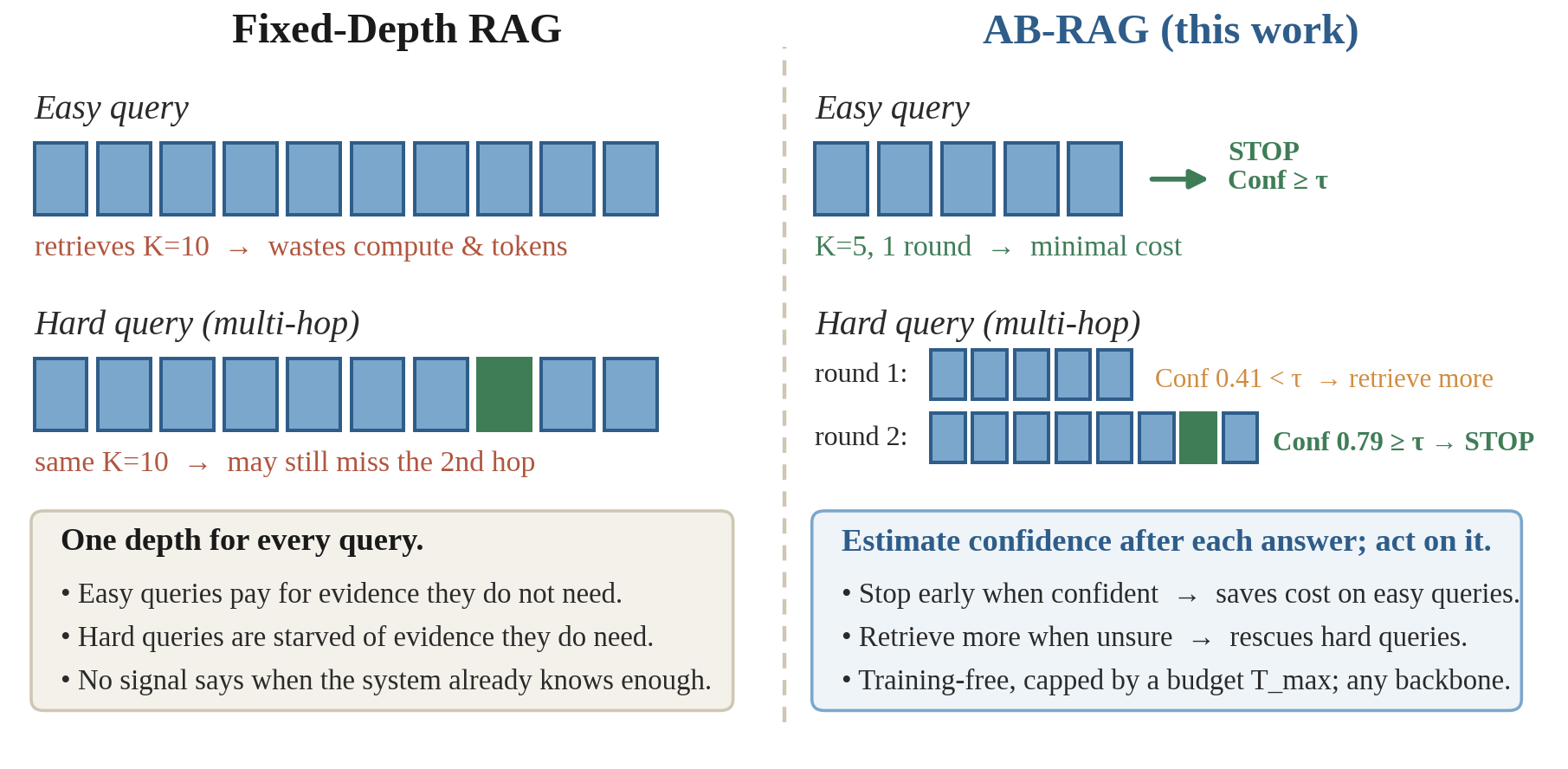}
\caption{Fixed-depth RAG applies the same retrieval budget to every query, which
over-retrieves for easy questions and under-retrieves for hard ones. AB-RAG estimates
confidence after each answer and retrieves more only when needed, subject to a budget.}
\label{fig:concept}
\end{figure*}

This paper introduces AB-RAG (Adaptive Budgeted Retrieval-Augmented Generation), a
framework that makes retrieval depth adaptive and budgeted. Rather than retrieving a
fixed number of passages, AB-RAG generates an answer, estimates how confident it is in
that answer, and then decides whether to stop or to retrieve more evidence. It repeats
this loop until the answer is confident enough or until a retrieval budget runs out. The
guiding principle is that a system should retrieve as much evidence as a question
actually needs, and no more. This yields three benefits at once: efficiency on easy
queries by stopping early, robustness on hard queries by retrieving more when unsure,
and a single tunable threshold that trades retrieval cost against accuracy.
Fig.~\ref{fig:concept} contrasts the two regimes.

An important property of AB-RAG is that it is training-free and backbone-agnostic. It
needs no fine-tuning, no separate reward model, and no special tokens, so it can wrap
around any existing generator. This sets it apart from earlier adaptive methods.
Self-RAG~\cite{asai2024selfrag}, for example, trains a model to emit reflection tokens,
and FLARE~\cite{jiang2023flare} relies on a single confidence signal with no explicit
retrieval budget. AB-RAG instead works with any backbone, whether it is an open-weight
model that exposes token-level probabilities or a closed commercial API that does not.
For closed APIs it substitutes a self-consistency proxy in place of the missing
probabilities. This matters because a large share of deployed QA systems today are built
on top of proprietary APIs, and a method that only works when internal probabilities are
available would not apply to them. The same backbone-agnostic stance guided our earlier
work on adaptive decision-making in multi-agent systems~\cite{kamthan2025marl}, where
agents act on their own local signals rather than a centrally trained controller.

\subsection{Problem Statement and Objectives}
The problem this work addresses is the unreliable and inefficient use of retrieval in
question answering. Fixed-depth RAG cannot match its retrieval effort to the difficulty
of a query, and a standard pipeline gives no signal about when a generated answer can be
trusted. The objectives of the project are as follows.
\begin{enumerate}
\item To design a training-free, confidence-driven adaptive retrieval framework that
adjusts retrieval depth for each query under an explicit budget.
\item To build a multi-signal confidence estimator that combines the model's internal
certainty, the agreement between the answer and the evidence, and the quality of the
retrieval, without training any model.
\item To make the framework operate on both open-weight backbones, which provide real
token log-probabilities, and closed APIs, which do not, using a self-consistency proxy
in the closed case.
\item To find out empirically which confidence signals actually predict answer
correctness, treating the proposed signals as hypotheses to be tested rather than
assuming they all work.
\item To evaluate the framework carefully across different model scales and two
datasets, and to report negative or nuanced findings as genuine results rather than
hiding them.
\end{enumerate}

\subsection{Contributions}
This work makes the following contributions.
\begin{itemize}
\item A training-free, budgeted, multi-signal adaptive retrieval framework that operates
on both open-weight and closed-API backbones, filling a gap left open by prior
adaptive-RAG methods.
\item A confidence estimator that reliably separates correct from incorrect answers on
every backbone tested, reaching a 57.6\% versus 0\% Exact Match split on a factoid
dataset, which supports a strong selective-prediction use case.
\item An honest empirical study of the three proposed confidence signals, showing that
only the model-certainty signal is strongly predictive, that an evidence-consistency
signal is unsuitable for short-answer QA for a clear mechanistic reason, and that a
retrieval-variance signal had its sign backwards and was corrected through measurement.
\item A demonstration that the whole approach is reproducible on a single consumer
laptop with a 4~GB GPU and only a few dollars of API spend.
\end{itemize}

\subsection{Organisation of the Paper}
The rest of the paper is organised as follows. Section~\ref{sec:related} reviews the
background and related work, placing AB-RAG against standard, trained, and training-free
adaptive RAG. Section~\ref{sec:method} presents the methodology: the system
architecture, the retrieval stack with its governing equations, the three-signal
confidence estimator, and the adaptive budgeted loop as a formal algorithm.
Section~\ref{sec:impl} describes the implementation, including the development
environment, the datasets, the model backbones, and an analysis of API cost.
Section~\ref{sec:results} reports all experimental results, covering retrieval quality,
the static-versus-adaptive comparison across three backbones, the confidence-correctness
analysis, the cost-accuracy tradeoff, and the signal ablation and diagnostics.
Section~\ref{sec:conclusion} concludes and sets out directions for future work.

\section{Background and Related Work}
\label{sec:related}

\subsection{Conceptual Background}
This section explains the concepts AB-RAG builds on, so the rest of the paper can be
read without assuming prior familiarity with retrieval systems or language-model
confidence.

\textit{Retrieval-Augmented Generation.}
A language model trained on a fixed corpus can only answer from what it memorised, and
it has no way to cite a source or stay current. RAG addresses this by adding a retrieval
step in front of generation. When a question arrives, a retriever searches a corpus for
passages likely to contain the answer, and those passages are placed in the model's
prompt as evidence. The model then answers using the evidence rather than its memory
alone, which reduces hallucination and lets the system work over knowledge it was never
trained on~\cite{shuster2021retrieval, gao2023survey}.

\textit{Sparse and dense retrieval.}
There are two main families of retriever. Sparse retrieval, represented here by BM25,
matches the literal words of the query against the words of each passage and scores
passages by term overlap, weighted so that rare words count for more and very long
passages are not unfairly favoured. It is fast and strong when the answer shares
vocabulary with the question, but it misses passages that express the same idea in
different words. Dense retrieval instead encodes the query and each passage into vectors
using a neural embedding model and measures similarity by the closeness of those
vectors. Because the vectors capture meaning rather than exact words, dense retrieval
can match paraphrases, but it depends heavily on the quality of the embedding model. A
common strategy is to combine the two so that the lexical precision of BM25 and the
semantic recall of dense retrieval reinforce each other.

\textit{Reranking.}
Retrievers like BM25 and dense search score each passage independently of the others and
are tuned for speed over a large corpus, so their ordering is only approximate. A
reranker improves the ordering of a small candidate set. A cross-encoder reranker takes
the query and a single passage together as one input and outputs a relevance score,
which lets it model fine-grained interactions that a fast retriever cannot. It is too
slow to run over a whole corpus, so it is applied only to the top candidates the
retriever returns.

\textit{Confidence and selective prediction.}
A model can produce an answer, but on its own it does not tell us whether that answer is
trustworthy. Confidence estimation tries to attach a number to an answer that reflects
how likely it is to be correct~\cite{guo2017calibration, kadavath2022know}. If that
number is reliable, it enables selective prediction, where the system answers when
confident and abstains or gathers more information when not. AB-RAG uses confidence in
exactly this way, as the signal that decides whether to stop or to retrieve more.

\subsection{Related Work}
\textit{Standard RAG.}
The retrieve-then-generate pattern was established by early RAG work~\cite{lewis2020rag}
and by Dense Passage Retrieval (DPR)~\cite{karpukhin2020dpr}, which showed that learned
dense retrievers could outperform traditional sparse methods on open-domain question
answering. These systems use a fixed retrieval depth and do not adapt to the query.
Later work such as REALM~\cite{guu2020realm}, RETRO~\cite{borgeaud2022retro}, and
Fusion-in-Decoder~\cite{izacard2021fid} scaled or restructured the retrieve-then-read
pipeline but kept retrieval essentially static with respect to per-query difficulty.
In-context retrieval~\cite{ram2023incontext}, nearest-neighbour language
models~\cite{he2021efficient}, tool-using models~\cite{schick2023toolformer,
nakano2021webgpt}, and compositional retrieve-and-reason
programs~\cite{khattab2022demonstrate, press2023measuring} extend the paradigm in other
directions, and standard retrieval benchmarks such as BEIR~\cite{thakur2021beir}, Natural
Questions~\cite{kwiatkowski2019natural}, and ExpertQA~\cite{malaviya2023expertqa} support
their evaluation. Language models are also known to encode substantial factual knowledge
directly in their parameters~\cite{petroni2019lama}, and structured-output settings
benefit from retrieval grounding as well~\cite{bechard2024reducing}. Retrieval has been
combined with few-shot learning at scale~\cite{izacard2022atlas} and with explicit
reasoning chains~\cite{wei2022chain} and reasoning-acting loops~\cite{yao2023react}, while
recent work has also focused on systematically evaluating RAG
systems~\cite{es2024ragas, chen2024benchmarking}.

\textit{Self-RAG.}
Self-RAG~\cite{asai2024selfrag} makes generation adaptive by training a model to emit
special reflection tokens that decide when to retrieve and that critique the retrieved
evidence and the generated answer. It is a strong method, but it requires training the
model with a specially constructed dataset, which makes it costly to apply and impossible
to use directly with a closed API whose weights cannot be modified.

\textit{FLARE.}
Forward-Looking Active Retrieval (FLARE)~\cite{jiang2023flare} decides when to retrieve
during long-form generation by watching the model's token probabilities and retrieving
more when an upcoming sentence looks uncertain. It is training-free, which AB-RAG shares,
but it relies on a single signal, token probability, and it sets no explicit retrieval
budget, so it cannot bound the cost of answering a query. Related iterative approaches
interleave retrieval with reasoning~\cite{trivedi2023interleaving} or repeatedly refine
the query~\cite{shao2023iterative}, but likewise without an explicit budget. A concurrent line of work makes retrieval
adaptive by routing queries according to predicted complexity~\cite{jeong2024adaptive} or
by detecting low-confidence spans during generation and validating
them~\cite{varshney2023stitch}, which is close in spirit to AB-RAG but either requires a
trained router or targets long-form generation rather than budgeted short-answer QA.

\textit{Retrieval components.}
The retrieval stack in this work uses established building blocks.
BM25~\cite{robertson2009bm25, robertson1995okapi} is the standard sparse ranking
function. BGE~\cite{xiao2024cpack} is a strong open-weight text embedding model used for
dense retrieval. Reciprocal Rank Fusion (RRF)~\cite{cormack2009rrf} is a simple and
robust way to combine the rankings of different retrievers without tuning. Cross-encoder
rerankers~\cite{nogueira2019bert} trained on the MS MARCO passage ranking
data~\cite{nguyen2016msmarco} are widely used to refine candidate ordering, and
ColBERT~\cite{khattab2020colbert} is a related late-interaction approach.

\textit{Confidence without training.}
For models that expose token probabilities, the average probability of the generated
tokens is a natural confidence signal~\cite{kadavath2022know, jiang2021calibration}. For
models that do not, self-consistency~\cite{wang2023selfconsistency} offers an
alternative: the model is sampled several times and the agreement among the samples is
used as a proxy for confidence, on the reasoning that a model which keeps giving the same
answer is more likely to be right. Calibration studies~\cite{guo2017calibration,
gal2016dropout} and analyses of when models know what they know~\cite{kadavath2022know,
mallen2023trust} motivate treating such signals as hypotheses to be measured rather than
trusted by default. AB-RAG uses the self-consistency idea to extend confidence estimation
to closed APIs. Other approaches ask the model to verbalise its own
uncertainty~\cite{lin2022teaching, si2023prompting} or estimate semantic uncertainty over
sampled generations~\cite{kuhn2023semantic}; these are complementary to the
self-consistency proxy used here.

\subsection{Outcome of the Review}
The review shows a clear gap. Standard RAG is training-free and works on any backbone but
is not adaptive. Self-RAG is adaptive but needs training and therefore cannot be applied
to closed APIs. FLARE is training-free and adaptive but uses a single signal and sets no
budget. No existing method combines all of the properties that a practical, reliable QA
system would want at the same time: a training-free design, an explicit retrieval budget,
a confidence estimate built from more than one signal, and operation on both open and
closed backbones. AB-RAG is designed to fill exactly this gap, and Table~\ref{tab:pos}
summarises the comparison.

\begin{table*}[t]
\centering
\caption{Positioning of AB-RAG Against Related RAG Methods}
\label{tab:pos}
\renewcommand{\arraystretch}{1.2}
\begin{tabular}{lcccc}
\toprule
\textbf{Method} & \textbf{Training-free} & \textbf{Budgeted} & \textbf{Multi-signal confidence} & \textbf{Works on closed APIs} \\
\midrule
Standard RAG~\cite{lewis2020rag} & Yes & No & No & Yes \\
Self-RAG~\cite{asai2024selfrag} & No & No & Reflection tokens & No \\
FLARE~\cite{jiang2023flare} & Yes & No & Single signal & Partial \\
\textbf{AB-RAG (ours)} & \textbf{Yes} & \textbf{Yes} & \textbf{Three signals} & \textbf{Yes} \\
\bottomrule
\end{tabular}
\end{table*}

\subsection{Technologies Used}
The implementation uses Python as the primary language. PyTorch~\cite{paszke2019pytorch}
provides the deep-learning runtime and the Hugging Face
libraries~\cite{wolf2020transformers} supply the language and embedding models.
FAISS~\cite{johnson2021faiss} indexes the dense vectors and searches them efficiently.
The \texttt{rank\_bm25} library provides the sparse retriever, and the
sentence-transformers library~\cite{reimers2019sbert} provides both the BGE embedding
model and the cross-encoder reranker. Open-weight models run locally through Hugging Face
Transformers and through Ollama, while the closed model is accessed through a commercial
API. All experiments run on a single consumer laptop GPU.

\section{Methodology and Framework}
\label{sec:method}

\subsection{Overall Framework}
The aim of AB-RAG is to answer a question using only as much retrieval as that question
needs, and to do so without training any model. The framework has three moving parts
that work together. First, a retrieval stack finds and orders candidate evidence for a
question. Second, a generator produces an answer from the current evidence. Third, a
confidence estimator scores how trustworthy that answer is, and a control loop uses the
score to decide whether to stop or to retrieve more. The loop is bounded by a budget, so
the system can never spend more than a fixed number of retrieval rounds on any single
question.

\subsection{System Architecture}
\begin{figure*}[t]
\centering
\includegraphics[width=0.95\textwidth]{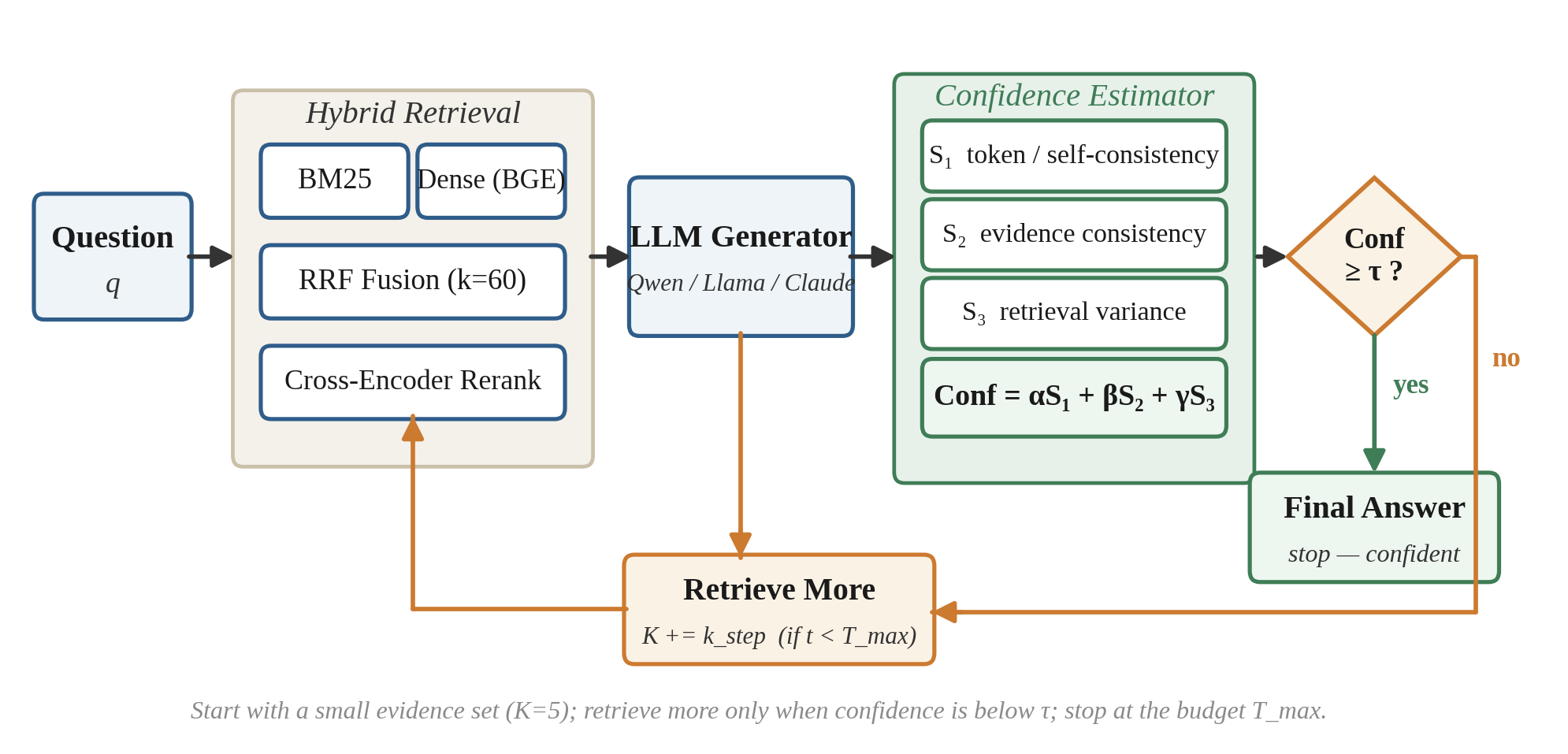}
\caption{The AB-RAG architecture. Hybrid retrieval and reranking produce an evidence
set; the generator answers; the confidence estimator combines three signals; and the
decision step either stops or triggers another retrieval round, subject to the budget.}
\label{fig:arch}
\end{figure*}

Fig.~\ref{fig:arch} shows the complete pipeline. A question enters the hybrid retrieval
stage, where a sparse retriever and a dense retriever each rank the corpus and their
rankings are fused. A cross-encoder then reranks the fused candidates so that the most
relevant passages rise to the top. The top passages become the evidence set, which is
passed to the generator together with the question. The generator returns an answer and
a first confidence signal. The confidence estimator combines that signal with two further
signals computed from the evidence and the retrieval scores, and produces a single
confidence value. A decision step compares this value against a threshold. If the answer
is confident enough the loop stops and returns it; if not, and if the budget has not been
spent, the system enlarges the evidence set and repeats the generate-and-score step. If
the budget is exhausted the loop stops and returns the best answer it has.

A key design choice is that the same architecture works for any generator. The only part
that depends on the backbone is the first confidence signal, which is read directly from
token probabilities when the model exposes them and is otherwise estimated by sampling.
Everything else, including the retrieval stack, the other two signals, and the control
loop, is identical across backbones.

\subsection{Retrieval Stack}
The retrieval stack combines a sparse retriever, a dense retriever, a fusion step, and a
reranker.

\textit{Sparse retrieval with BM25.}
BM25~\cite{robertson2009bm25} scores a passage against a query by summing a weight for
each query term that appears in the passage. The weight rewards terms that are rare
across the corpus and discounts terms that appear very often within a single passage,
with a correction for passage length. For a query $q$ and passage $d$, the score is
\begin{equation}
\text{BM25}(q,d) = \sum_{t \in q} \text{IDF}(t) \cdot
\frac{f(t,d)\,(k_1 + 1)}{f(t,d) + k_1\!\left(1 - b + b\,\dfrac{|d|}{\text{avgdl}}\right)}
\end{equation}
where $f(t,d)$ is the frequency of term $t$ in passage $d$, $|d|$ is the passage length,
$\text{avgdl}$ is the average passage length, $\text{IDF}(t)$ is the inverse document
frequency of $t$, and $k_1$ and $b$ control term-frequency saturation and length
normalisation. We use the standard values $k_1 = 1.5$ and $b = 0.75$.

\textit{Dense retrieval.}
The dense retriever encodes the query and each passage into unit-length vectors with the
BGE embedding model~\cite{xiao2024cpack} and scores a passage by the cosine similarity
between its vector and the query vector. Because the vectors are normalised, the cosine
similarity is the dot product,
\begin{equation}
\text{sim}(q,d) = e(q) \cdot e(d)
\end{equation}
where $e(q)$ and $e(d)$ are the embeddings of the query and passage. The vectors are
indexed with FAISS~\cite{johnson2021faiss} so the most similar passages can be found
quickly even over a large corpus.

\textit{Fusion with Reciprocal Rank Fusion.}
The sparse and dense rankings are combined using Reciprocal Rank
Fusion~\cite{cormack2009rrf}, which only needs the rank of each passage in each list and
not the raw scores, so it does not require the two retrievers to be on the same scale. A
passage at rank $r$ in a list contributes a score of one over a constant plus that rank,
and the contributions from the two lists are added,
\begin{equation}
\text{RRF}(d) = \sum_{i} \frac{1}{k + \text{rank}_i(d)}
\end{equation}
where the sum runs over the retrievers, $\text{rank}_i(d)$ is the rank of passage $d$ in
retriever $i$, and $k$ is a smoothing constant set to the standard value of 60. Passages
are then ordered by their fused score.

\textit{Reranking.}
The fused candidate list is reranked by a cross-encoder~\cite{nogueira2019bert}, which
scores each question--passage pair jointly. The cross-encoder is more accurate than the
retrievers but slower, so it is applied only to the candidate pool rather than the whole
corpus. The reranked order defines the final evidence ranking, and the top passages from
it form the evidence set given to the generator.

\subsection{The Confidence Estimator}
\begin{figure*}[t]
\centering
\includegraphics[width=0.92\textwidth]{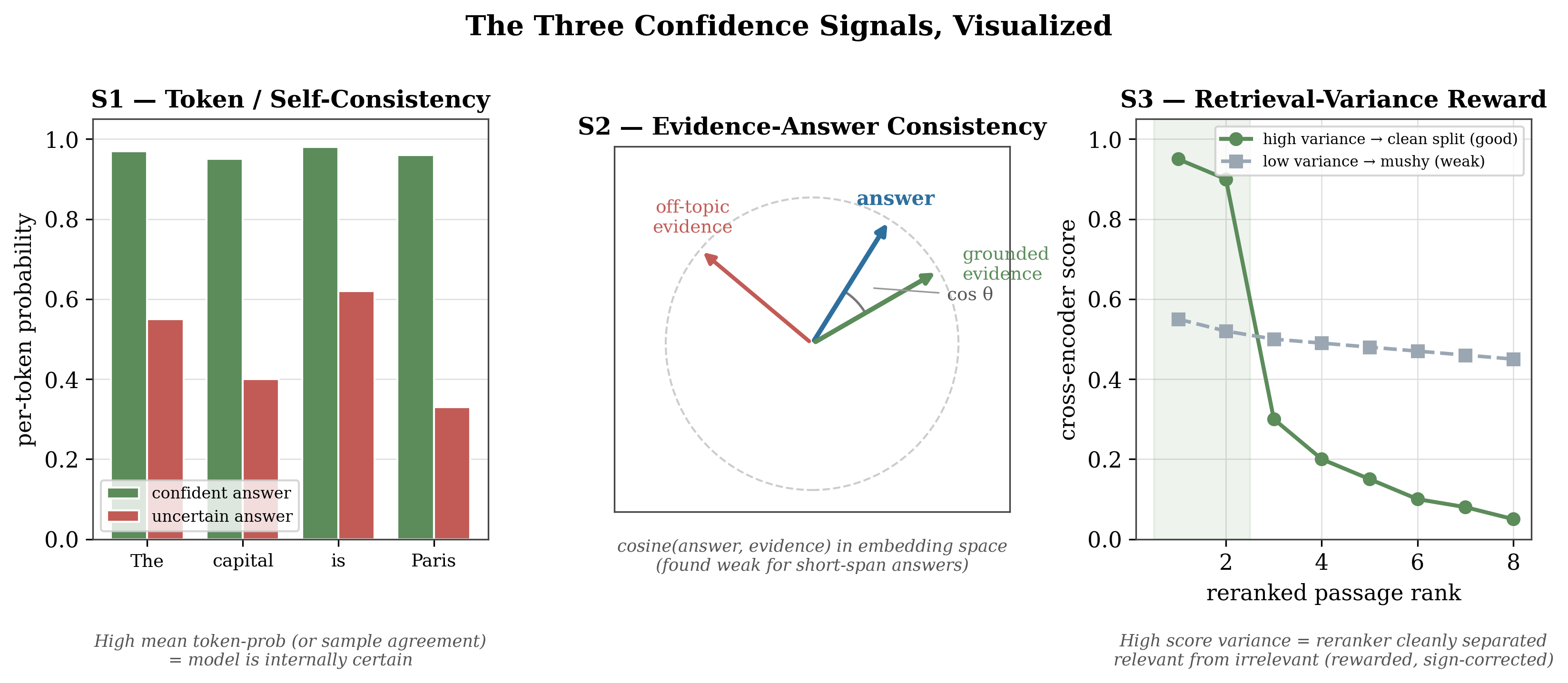}
\caption{The three confidence signals. $S_1$ is the model's own certainty from token
probabilities or self-consistency; $S_2$ is the embedding similarity between the answer
and the evidence; $S_3$ is the variance of the reranker scores, used as a reward for
clean separation.}
\label{fig:signals}
\end{figure*}

The confidence estimator is the core contribution of this work. After the generator
produces an answer, the estimator combines three signals into a single confidence value,
\begin{equation}
\text{Conf} = \text{clip}\!\left(\alpha S_1 + \beta S_2 + \gamma S_3,\; 0,\; 1\right)
\end{equation}
where $S_1$, $S_2$, and $S_3$ are the three signals described below and $\alpha, \beta,
\gamma$ are weights. The value is clipped to $[0,1]$ so it can be read as a confidence.
The three signals capture three different and complementary notions of trust: how sure
the model is, how well the answer agrees with the evidence, and how cleanly the retrieval
separated relevant from irrelevant passages. Fig.~\ref{fig:signals} illustrates what each
signal measures.

\textit{Signal 1: token probability or self-consistency.}
The first signal measures the model's own certainty. For an open-weight model that
exposes token probabilities, it is the mean probability of the generated answer tokens,
\begin{equation}
S_1 = \frac{1}{N}\sum_{i=1}^{N} p(\text{token}_i)
\end{equation}
where $p(\text{token}_i)$ is the probability the model assigned to the $i$-th generated
token and $N$ is the number of tokens. A model that is internally certain assigns high
probability to each token it emits, giving a high signal. For a closed API that does not
expose token probabilities, the same notion is estimated by
self-consistency~\cite{wang2023selfconsistency}: the model is sampled $k$ times at a
non-zero temperature, and the signal is the fraction of samples that agree with the most
common answer,
\begin{equation}
S_1 = \frac{\text{count of modal answer}}{k}.
\end{equation}
A model that keeps returning the same answer across samples is treated as more confident.
This substitution is what allows AB-RAG to run on closed APIs without any access to their
internals.

\textit{Signal 2: evidence--answer consistency.}
The second signal measures whether the answer is grounded in the retrieved evidence. It
is the cosine similarity between the embedding of the answer and the embedding of the
evidence, mapped to $[0,1]$,
\begin{equation}
S_2 = \frac{\cos\!\left(e(a), e(E)\right) + 1}{2}
\end{equation}
A high value means the answer is semantically close to the evidence, which is intended to
indicate grounding. As Section~\ref{sec:results} shows, this signal turns out not to
predict correctness for short-answer question answering, and the reason is examined there.

\textit{Signal 3: retrieval-score variance as a reward.}
The third signal is computed from the reranker scores of the evidence passages. It is the
variance of those scores, normalised to $[0,1]$,
\begin{equation}
S_3 = \text{Var}\!\left(\text{normalised rerank scores}\right)
\end{equation}
The interpretation deserves care. A high variance means the reranker assigned clearly
different scores to different passages, which indicates that it separated relevant
passages from irrelevant ones with confidence. A low variance means the scores are
bunched together, which indicates that the reranker could not tell the passages apart.
High variance is therefore a sign of good retrieval, so the signal is added as a reward
rather than subtracted as a penalty. The sign of this signal was corrected during the
diagnostics in Section~\ref{sec:results}, where the original penalty formulation was
found to be backwards.

The final weights are $\alpha = 0.7$, $\beta = 0.05$, and $\gamma = 0.25$. These give
most of the weight to the model's own certainty, a small weight to evidence consistency,
and a moderate weight to the retrieval-variance reward. The justification for these
values, including why the evidence-consistency weight is kept near zero, comes from the
signal ablation in Section~\ref{sec:results}, where the signals were treated as
hypotheses and tested rather than assumed to be useful.

\subsection{The Adaptive Budgeted Loop}
The control loop ties the pieces together. It begins with a small evidence set and
generates an answer, then computes the confidence. If the confidence reaches the
threshold the loop returns the answer immediately. If not, and if the budget of retrieval
rounds has not been spent, the loop enlarges the evidence set by a fixed step and repeats.
If the budget is spent the loop returns the best answer it has produced.
Algorithm~\ref{alg:abrag} states this precisely.

\begin{algorithm}[t]
\caption{AB-RAG --- Adaptive Budgeted Retrieval}
\label{alg:abrag}
\begin{algorithmic}[1]
\Require question $q$, corpus $C$, threshold $\tau$, budget $T_{\max}$,
start size $K_0$, step $k_{\text{step}}$
\State $t \gets 1$;\quad $K \gets K_0$;\quad $best \gets \varnothing$
\While{\textbf{true}}
  \State $E \gets \textsc{Rerank}(\textsc{Retrieve}(q, C))[1{:}K]$
  \State $(a, S_1) \gets \textsc{Generate}(q, E)$
  \State $S_2 \gets \textsc{EvidenceConsistency}(a, E)$
  \State $S_3 \gets \textsc{Var}(\textsc{RerankScores}(E))$
  \State $\textit{conf} \gets \text{clip}(\alpha S_1 + \beta S_2 + \gamma S_3, 0, 1)$
  \State $best \gets a$
  \If{$\textit{conf} \geq \tau$} \Return $a$ \Comment{confident: stop early}
  \EndIf
  \If{$t \geq T_{\max}$} \Return $best$ \Comment{budget spent: stop}
  \EndIf
  \State $t \gets t + 1$;\quad $K \gets K + k_{\text{step}}$
  \Comment{retrieve more}
\EndWhile
\end{algorithmic}
\end{algorithm}

The loop has three parameters: the confidence threshold $\tau$, which sets how sure the
system must be before it stops; the budget $T_{\max}$, which caps the number of retrieval
rounds; and the step $k_{\text{step}}$, which sets how many extra passages are added each
round. Their default values are $\tau = 0.6$, $T_{\max} = 3$, and $k_{\text{step}} = 5$,
with the evidence set starting at five passages. Raising $\tau$ makes the system more
cautious, so it retrieves more often and spends more, while lowering $\tau$ makes it stop
sooner and spend less. This single threshold is the knob that trades cost against
accuracy, and Section~\ref{sec:results} shows the resulting tradeoff curve.

\begin{figure}[t]
\centering
\includegraphics[width=0.86\columnwidth]{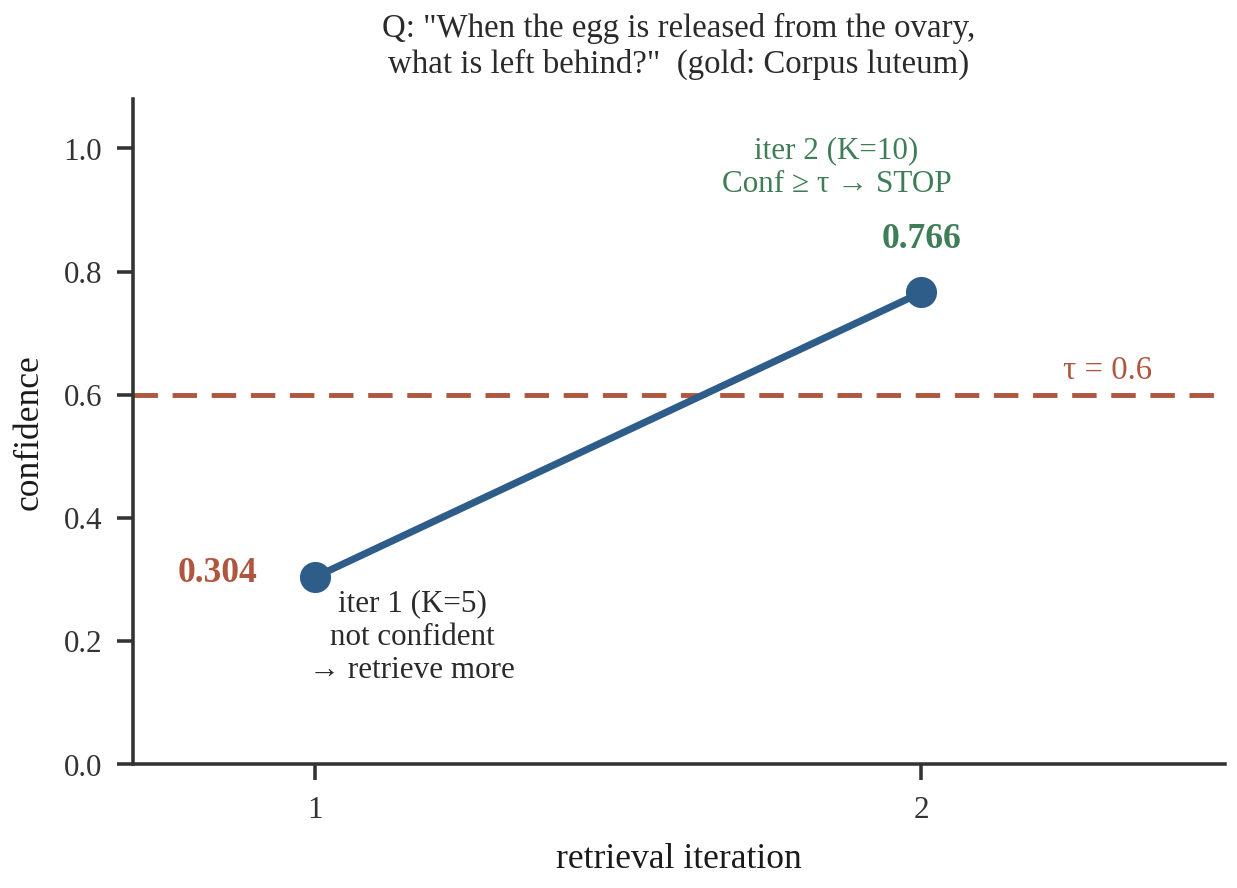}
\caption{A real worked example of the adaptive loop. The first round is below the
threshold and triggers more retrieval; the second round crosses the threshold and the
loop stops with the correct answer. The confidence values are taken from the actual run.}
\label{fig:worked}
\end{figure}

Fig.~\ref{fig:worked} shows the loop on a real question from the experiments. In the
first round the model answers from five passages but is not confident, so the loop
retrieves more. In the second round, with ten passages, the answer settles to the correct
form and the confidence rises above the threshold, so the loop stops. This is the
adaptive behaviour in miniature, using the actual confidence values recorded during the
run.

\subsection{Evaluation Metrics}
The framework is evaluated with four measures. Exact Match (EM) is the fraction of
answers that exactly match the gold answer after a normalisation that lowercases the text,
removes punctuation, and drops articles,
\begin{equation}
\text{EM} = \frac{\text{number of exact matches}}{\text{total questions}}
\end{equation}
Token-level F1 gives partial credit by measuring the overlap of words between the
predicted and gold answers, as the harmonic mean of precision and recall over shared
tokens,
\begin{equation}
\text{F1} = \frac{2 \cdot \text{precision} \cdot \text{recall}}{\text{precision} + \text{recall}}
\end{equation}
The Area Under the Receiver Operating Characteristic curve (AUROC) measures how well the
confidence value separates correct from incorrect answers. It can be read as the
probability that a randomly chosen correct answer receives a higher confidence than a
randomly chosen incorrect one,
\begin{equation}
\text{AUROC} = P(\text{Conf}_{\text{correct}} > \text{Conf}_{\text{incorrect}})
\end{equation}
An AUROC of one means the confidence perfectly ranks correct above incorrect answers,
while an AUROC of one half means the confidence is no better than chance. The fourth
measure is the average number of retrieval iterations per query, which captures the cost
side of the cost-accuracy tradeoff. Table~\ref{tab:hyper} lists all hyperparameters so the
configuration is reproducible.

\begin{table}[t]
\centering
\caption{Hyperparameters Used Across the AB-RAG Pipeline}
\label{tab:hyper}
\renewcommand{\arraystretch}{1.15}
\begin{tabular}{p{2.55cm}p{2.6cm}p{2.4cm}}
\toprule
\textbf{Component} & \textbf{Parameter} & \textbf{Value} \\
\midrule
Embedding model & encoder & bge-large-en-v1.5 \\
Reranker & cross-encoder & ms-marco-MiniLM-L12 \\
BM25 & $k_1$, $b$ & 1.5, 0.75 \\
Sparse/dense depth & bm25\_topk, dense\_topk & 20, 20 \\
Rank fusion & RRF $k$ & 60 \\
Reranking & rerank\_topk & 5 \\
Confidence weights & $\alpha$, $\beta$, $\gamma$ & 0.7, 0.05, 0.25 \\
Adaptive policy & $\tau$, $T_{\max}$, $k_{\text{step}}$ & 0.6, 3, 5 \\
Generation & max\_new\_tokens & 64 \\
Self-consistency & samples $k$ & 3 \\
Reproducibility & seed & 42 \\
\bottomrule
\end{tabular}
\end{table}

\section{Implementation}
\label{sec:impl}

\subsection{Development Environment}
All experiments were carried out on a single consumer laptop with an NVIDIA RTX~3050
Laptop GPU with 4~GB of video memory. The software environment used Python~3.11 inside a
Conda environment, with PyTorch built against CUDA~12.1. The retrieval and reranking
models, the local generator, and the embedding model all run on this one GPU, and the
closed model is reached over the network through a commercial API. Keeping the whole
project on a single modest machine was a deliberate constraint, because it shows that the
method does not depend on large compute and can be reproduced by a student or a small
team.

This constraint also shaped one important design decision about the retrieval corpus. The
most thorough way to evaluate open-domain retrieval would be to index the full Wikipedia
passage collection used by Dense Passage Retrieval~\cite{karpukhin2020dpr}, which contains
around twenty-one million passages. Indexing a corpus of that size is not feasible on a
4~GB laptop GPU and would take days. Instead, this work uses a pooled-corpus
open-retrieval setting, described below, which keeps the retrieval task genuinely
difficult while remaining tractable on the available hardware. This choice is stated
openly rather than hidden, because being honest about the scope of the evaluation is part
of the integrity of the work, and the pooled-corpus setting is itself a recognised and
citable way to study retrieval under controlled conditions.

\subsection{Datasets}
Two question answering datasets were used, chosen so the framework could be tested on two
different kinds of question.

HotpotQA~\cite{yang2018hotpotqa} is a multi-hop dataset, where answering a question
requires combining facts from more than one passage. It is used in two settings. In the
distractor setting, each question comes with about ten candidate passages, two of which
are the gold supporting passages, and the task is to find the two correct passages among
the ten. This setting is easy and saturates quickly, which made it useful mainly as a
first check of the retrieval code. In the open setting, the passages from many questions
are pooled into one shared corpus of several thousand passages, and each question must
find its two gold passages among the whole pool. This open setting is the main one used
here, because it keeps recall meaningful at every retrieval depth and lets reranking show
a real effect.

TriviaQA~\cite{joshi2017triviaqa} is a factoid dataset, where each question has a short
answer that can usually be found in a single passage. It is used in the open setting with
its own pooled corpus. TriviaQA was added as a second dataset to test whether the findings
from HotpotQA carry over to a different style of question. Table~\ref{tab:data} summarises
the datasets and their settings.

\begin{table}[t]
\centering
\caption{Datasets and Evaluation Settings}
\label{tab:data}
\renewcommand{\arraystretch}{1.2}
\begin{tabular}{llccl}
\toprule
\textbf{Dataset} & \textbf{Setting} & \textbf{Questions} & \textbf{Corpus} & \textbf{Answer type} \\
\midrule
HotpotQA & distractor & 500 & {\small$\sim$10/q} & multi-hop \\
HotpotQA & open (pooled) & 500 & 4{,}963 & multi-hop \\
TriviaQA & open (pooled) & 200 & 2{,}582 & factoid \\
\bottomrule
\end{tabular}
\end{table}

\subsection{Pipeline Modules}
The system is organised as a sequence of numbered scripts, each producing an output that
the next stage can reuse without recomputation. This structure was chosen so that
expensive steps, such as generating answers from a model, are run once and saved, and
later analysis reads the saved results rather than calling the model again. The adaptive
loop saves the full per-iteration trace for every question, recording the answer, the
confidence, and the three signal values at each step. Because these traces are saved, the
later experiment, ablation, and diagnostic scripts can replay the loop under different
settings without ever calling a model again, which makes the analysis both fast and free
to repeat.

The confidence signals are implemented as small, self-contained functions. The following
extract shows the token-probability and retrieval-variance signals as they appear in the
code.

\begin{figure}[t]
\begin{footnotesize}
\hrule\vspace{3pt}
\begin{verbatim}
def token_probability_confidence(token_logprobs):
    if not token_logprobs:
        return 0.0
    probs = np.exp(np.array(token_logprobs))
    return float(np.clip(probs.mean(), 0, 1))

def retrieval_score_variance(rerank_scores):
    s = np.array(rerank_scores, dtype=np.float64)
    if s.size <= 1:
        return 0.0
    lo, hi = s.min(), s.max()
    if hi - lo < 1e-9:
        return 0.0
    s_norm = (s - lo) / (hi - lo)
    return float(np.clip(s_norm.var(), 0, 1))
\end{verbatim}
\vspace{1pt}\hrule
\end{footnotesize}
\caption{The token-probability and retrieval-variance confidence signals as implemented.
The variance signal normalises the reranker scores before measuring their spread.}
\label{fig:code}
\end{figure}

The generation module is where the difference between open and closed backbones is
handled. For the local open-weight model, the code reads the real per-token
log-probabilities the model returns. For the closed model and the Ollama-hosted model,
which do not return token probabilities, the code instead samples the model several times
and measures how often the samples agree, using that agreement as the Signal~1 proxy.

\subsection{Model Backbones}
Three generators were used so the framework could be tested across a range of model sizes
and both open and closed access.

The first is Qwen2.5-1.5B-Instruct~\cite{yang2024qwen}, a small open-weight model that
runs locally and returns real token probabilities. It fits within the 4~GB of GPU memory
in half precision and represents the low end of the capability range. The second is
Llama-3.2-3B~\cite{dubey2024llama}, a mid-sized open-weight model accessed through Ollama,
which does not expose token probabilities, so it uses the self-consistency proxy. The
third is Claude Haiku~\cite{anthropic2024claude}, a closed commercial model accessed
through a hosted API, which also does not expose token probabilities and so also uses
self-consistency. Using these three together is what makes the cross-backbone analysis in
Section~\ref{sec:results} possible: the small local model shows what happens at low
capability, the mid model shows the open-weight middle of the range, and the closed API
model shows the practical case that motivated the self-consistency design in the first
place.

\subsection{API Cost and Token Analysis}
Because one backbone is a paid API, the cost of running AB-RAG on it is worth examining,
both as a practical matter and because it interacts with the design. The self-consistency
proxy requires sampling the model several times for each answer, with three samples used
here, so each generated answer costs three API calls rather than one. This is the price of
estimating confidence without access to token probabilities.

At the time of the experiments, the closed model was priced at one dollar per million
input tokens and five dollars per million output tokens. A static run over two hundred
questions, which generates one answer per question with three samples each, makes about
six hundred API calls. Each call sends the question and the evidence passages and receives
a short answer, so the input dominates the cost. The adaptive runs make more calls because
some questions trigger additional retrieval rounds, but the budget caps how far this can
grow. Table~\ref{tab:cost} sets out the estimated cost of each Claude experiment.

\begin{table}[t]
\centering
\caption{Estimated API Cost of the Claude Haiku Experiments}
\label{tab:cost}
\renewcommand{\arraystretch}{1.2}
\begin{tabular}{lccc}
\toprule
\textbf{Experiment} & \textbf{API calls} & \textbf{Tokens (in/out)} & \textbf{Est. cost} \\
\midrule
HotpotQA static (K=10) & 600 & $\sim$1{,}340 / 20 & $\sim$\$0.86 \\
TriviaQA static (K=10) & 600 & $\sim$1{,}340 / 20 & $\sim$\$0.86 \\
HotpotQA adaptive & $\sim$816 & $\sim$1{,}080 / 20 & $\sim$\$0.96 \\
TriviaQA adaptive & $\sim$1{,}086 & $\sim$1{,}210 / 20 & $\sim$\$1.42 \\
\bottomrule
\end{tabular}
\\[2pt]
\begin{minipage}{\columnwidth}
\footnotesize Priced at \$1 per million input tokens and \$5 per million output tokens,
with three self-consistency samples per answer.
\end{minipage}
\end{table}

Two points stand out. First, the whole study, across all three backbones and two datasets,
was completed on a single laptop with only a few dollars of API spend, which supports the
claim that the method is reproducible on a modest budget. Second, the budget in the
adaptive loop is not only an accuracy control but also a direct cost control, because it
bounds the number of API calls a single question can ever cause. This makes AB-RAG
attractive in settings where each model call has a real monetary cost.

\section{Results and Analysis}
\label{sec:results}

\subsection{Retrieval Results}
The first set of experiments measures how well the retrieval stack finds the gold
passages, because the quality of the evidence sets an upper bound on how well the
generator can answer. Recall at $k$, the fraction of gold passages found within the top
$k$ retrieved passages, is reported for the sparse retriever (BM25), the dense retriever,
their hybrid fusion, and the reranked hybrid.

\textit{Distractor setting.}
The first measurements were taken in the HotpotQA distractor setting, where each question
has only about ten candidate passages. Recall reaches one hundred percent at $k=10$ and
$20$, which is expected because there are only about ten passages per question, so
retrieving ten of them finds everything. Only recall at five separates the methods, where
the dense retriever leads (92.9\%), followed by the hybrid (88.2\%) and then BM25
(73.6\%). This saturation is why the distractor setting was used only as an early check:
when almost every passage is retrieved regardless of method, the numbers cannot show
whether one retriever is genuinely better, and adding the cross-encoder leaves recall
almost unchanged (87.5\% versus 88.2\% at five) because there is nothing for the reranker
to fix.

\begin{figure*}[t]
\centering
\includegraphics[width=0.96\textwidth]{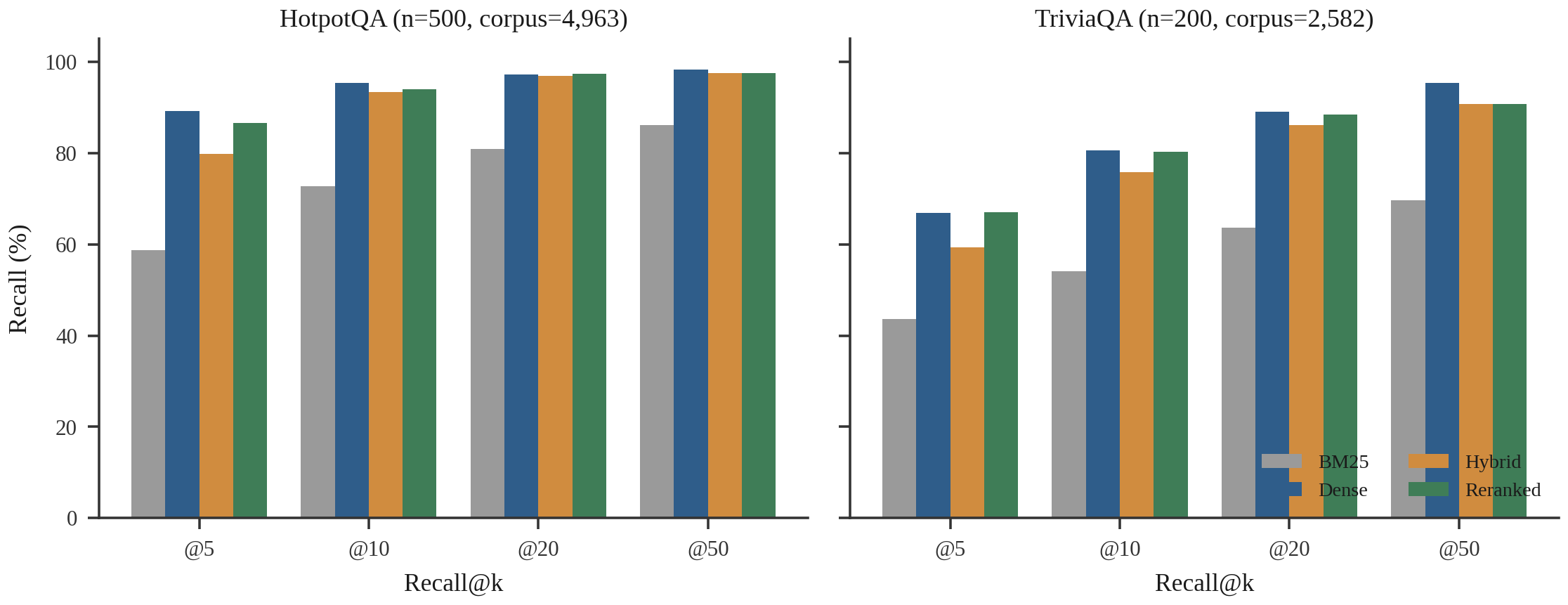}
\caption{Open-retrieval recall by method on HotpotQA and TriviaQA. The dense retriever is
strongest on both datasets, hybrid fusion sits below it on this clean text, and reranking
lifts the hard low-$k$ recall.}
\label{fig:recall}
\end{figure*}

\textit{Open setting.}
Moving to the open setting, where each question must find its gold passages among several
thousand pooled passages, makes the task genuinely hard and the numbers informative.
Table~\ref{tab:recall} shows recall on both open corpora, and Fig.~\ref{fig:recall} plots
them side by side.

\begin{table}[t]
\centering
\caption{Retrieval Recall (\%) in the Open Setting}
\label{tab:recall}
\renewcommand{\arraystretch}{1.15}
\begin{tabular}{lcccc}
\toprule
\textbf{Method} & \textbf{@5} & \textbf{@10} & \textbf{@20} & \textbf{@50} \\
\midrule
\multicolumn{5}{l}{\textit{HotpotQA (n=500, corpus 4{,}963)}} \\
BM25 & 58.7 & 72.8 & 80.9 & 86.2 \\
Dense & 89.2 & 95.3 & 97.2 & 98.3 \\
Hybrid & 79.8 & 93.3 & 96.9 & 97.6 \\
Reranked & 86.6 & 94.0 & 97.3 & 97.6 \\
\midrule
\multicolumn{5}{l}{\textit{TriviaQA (n=200, corpus 2{,}582)}} \\
BM25 & 43.6 & 54.1 & 63.7 & 69.6 \\
Dense & 66.9 & 80.6 & 89.1 & 95.3 \\
Hybrid & 59.4 & 75.8 & 86.2 & 90.8 \\
Reranked & 67.0 & 80.3 & 88.4 & 90.8 \\
\bottomrule
\end{tabular}
\end{table}

Three findings come out of the open-setting results, and they hold on both datasets.
First, recall no longer saturates, so the numbers are meaningful at every depth, which
confirms that the open setting is the right one for studying retrieval. Second, the dense
retriever is the single strongest method on this clean encyclopaedic text. This is worth
stating plainly because it runs against the common expectation that hybrid retrieval
always wins. The explanation is that when the dense retriever is already strong and the
text is clean, fusing in the lexical matches from BM25 pulls in passages that share words
with the query but are not actually relevant, which slightly lowers precision at the top
of the ranking. Hybrid retrieval helps more when the dense retriever is weak or the text
is noisy, neither of which is the case here. This is reported as an honest nuance rather
than smoothed over, because it reflects what the data actually shows. Third, reranking
helps in the open setting, lifting hybrid recall at five from 79.8 to 86.6 on HotpotQA,
the opposite of what happened in the saturated distractor setting, which confirms that the
cross-encoder earns its place when there are real candidates to reorder.

\subsection{Static RAG versus AB-RAG Across Backbones}
\begin{figure*}[t]
\centering
\includegraphics[width=0.92\textwidth]{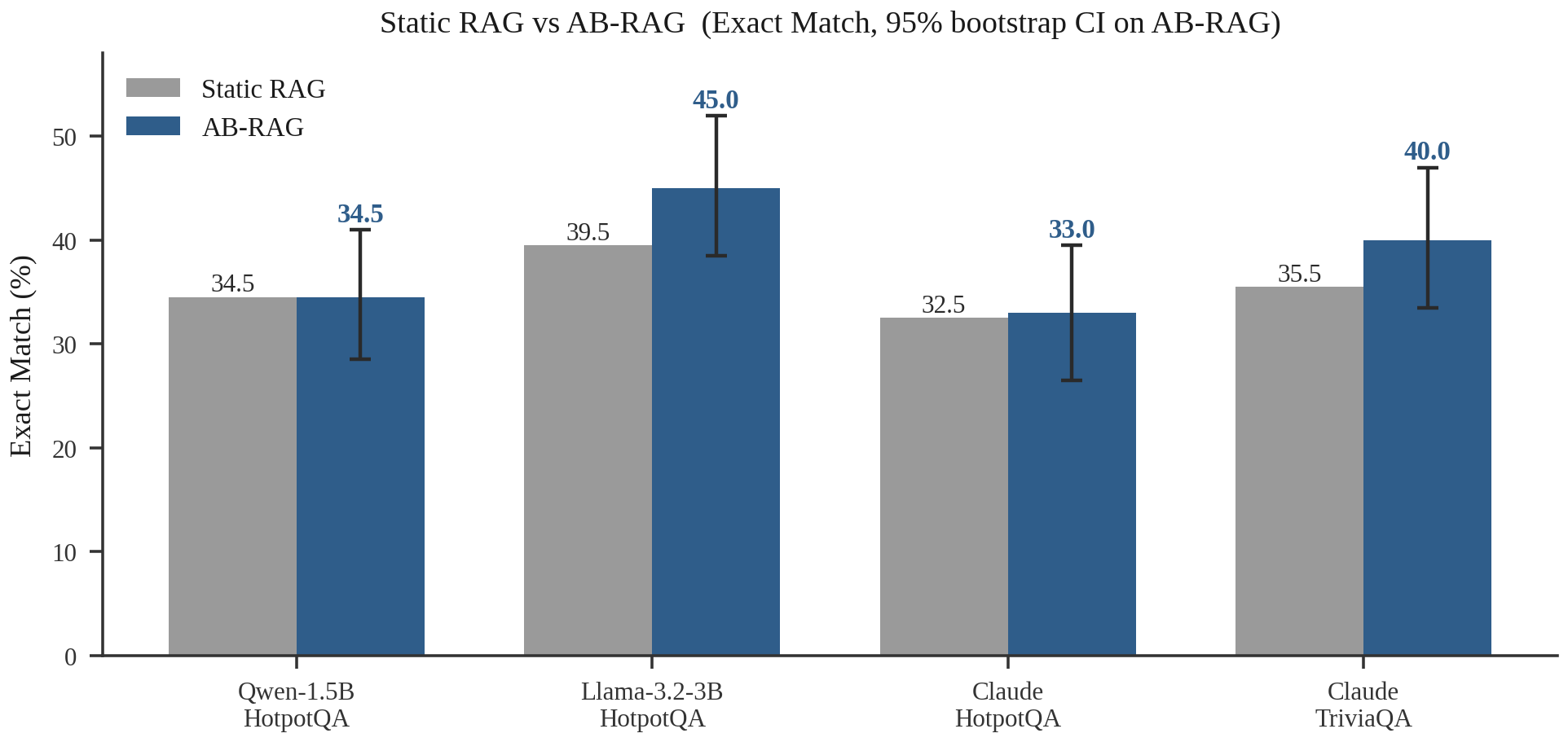}
\caption{Exact Match for static RAG and AB-RAG across backbones, with 95\% bootstrap
confidence intervals on the AB-RAG values. AB-RAG improves over static most clearly on the
mid-sized model and on the closed model with TriviaQA.}
\label{fig:static}
\end{figure*}

The central experiments compare static RAG, which uses a fixed retrieval depth and a
single pass, against AB-RAG, which adapts the retrieval depth per query. The comparison
was run on three backbones of increasing capability, and on both datasets for the closed
model. Table~\ref{tab:static} reports Exact Match, token-level F1, the average number of
retrieval iterations, and the AUROC of confidence against correctness, with bootstrap
confidence intervals on the AB-RAG Exact Match. Fig.~\ref{fig:static} shows the Exact
Match results.

\begin{table*}[t]
\centering
\caption{Static RAG versus AB-RAG Across Backbones}
\label{tab:static}
\renewcommand{\arraystretch}{1.2}
\begin{tabular}{lcccccc}
\toprule
\textbf{Backbone / Dataset} & \textbf{Static EM} & \textbf{AB-RAG EM} & \textbf{Static F1} & \textbf{AB-RAG F1} & \textbf{Avg. iter.} & \textbf{AUROC} \\
\midrule
Qwen-1.5B / HotpotQA & 34.5 & 34.5 & 47.3 & 46.2 & 1.09 & 0.600 \\
Llama-3.2-3B / HotpotQA & 39.5 & \textbf{45.0} & 56.3 & 58.3 & 1.90 & 0.644 \\
Claude Haiku / HotpotQA & 32.5 & 33.0 & 51.5 & 50.7 & 1.36 & 0.682 \\
Claude Haiku / TriviaQA & 35.5 & \textbf{40.0} & 50.8 & 53.5 & 1.81 & 0.733 \\
\bottomrule
\end{tabular}
\\[2pt]
\begin{minipage}{0.9\textwidth}
\centering
\footnotesize EM and F1 are percentages; Avg.\ iter.\ is the average retrieval iterations;
AUROC is for confidence against correctness. Bootstrap 95\% CIs on AB-RAG EM: Qwen
28.5--41.0, Llama 38.5--52.0, Claude/HotpotQA 26.5--39.5, Claude/TriviaQA 33.5--47.0.
\end{minipage}
\end{table*}

The results show a clear pattern that depends on the capability of the backbone. With the
small Qwen-1.5B model, AB-RAG and static RAG tie on Exact Match, and the average number of
iterations barely rises above one, which means the loop almost always stops after the
first round. A small model rarely improves its answer when given more evidence, so the
adaptive policy has little to work with. This is reported as an honest null result, not
hidden, because it marks the lower edge of where the method helps.

With the mid-sized Llama-3.2-3B model, AB-RAG improves Exact Match over static from 39.5
to 45.0, and the average number of iterations rises to 1.90, showing that the loop is
genuinely engaging the adaptive behaviour. With the closed Claude Haiku model on HotpotQA
the two methods are close on Exact Match, but on TriviaQA AB-RAG improves Exact Match over
static from 35.5 to 40.0. The TriviaQA result is notable because it shows the accuracy
gain that the multi-hop HotpotQA setting did not produce for this model. Factoid
questions, which often have a single findable answer, benefit more from the strategy of
retrieving more when unsure than multi-hop questions do here.

One number in Table~\ref{tab:static} needs care to interpret. The Llama model's Exact
Match of 45.0 is higher than the Claude model's 33.0 on HotpotQA, but this should not be
read as Llama being a better question answering model than Claude. Exact Match requires
the predicted answer to match the gold answer as an exact string after normalisation, and
Claude tends to give slightly more verbose answers, such as ``Wichita, Kansas'' when the
gold answer is simply ``Wichita''. These answers are correct in substance but are marked
wrong by Exact Match. The token-level F1 scores, which give partial credit, narrow the
gap, and a set of qualitative examples in Section~\ref{sec:results} shows this effect
directly. The honest reading is that AB-RAG behaves consistently across backbones, and
that Exact Match alone understates the closed model because of answer formatting.

\subsection{Confidence Predicts Correctness}
\begin{figure*}[t]
\centering
\includegraphics[width=0.92\textwidth]{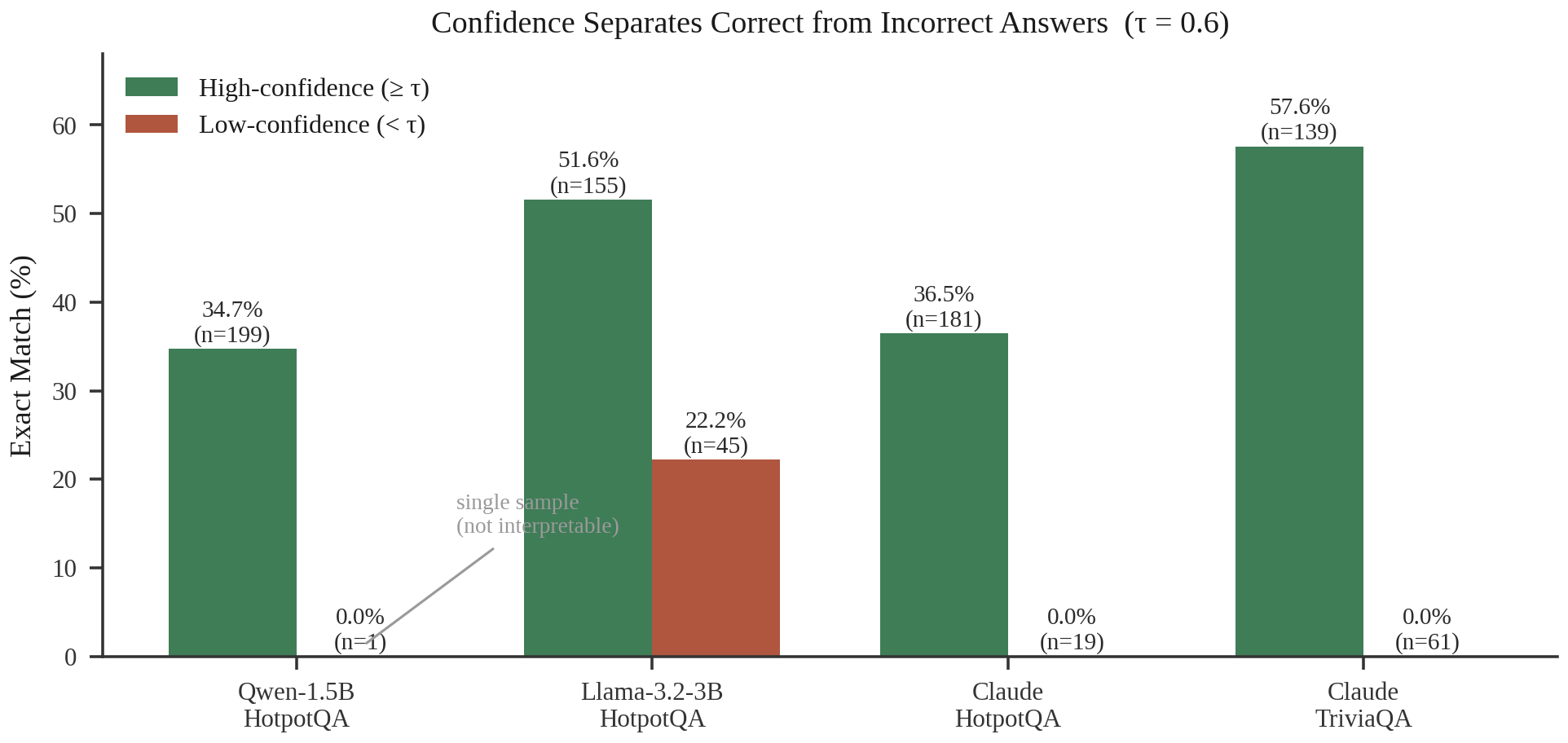}
\caption{High-confidence answers achieve far higher Exact Match than low-confidence
answers on every backbone and dataset. The closed model on TriviaQA shows the cleanest
separation, 57.6\% against zero, with a large low-confidence group.}
\label{fig:confsep}
\end{figure*}

The most important claim of this work is that the confidence value produced by the
estimator actually predicts whether an answer is correct. If it does, then the confidence
can be trusted to decide when to stop retrieving and when to abstain. This was tested by
splitting the answers into a high-confidence group and a low-confidence group at the
threshold, and measuring the Exact Match of each group. Table~\ref{tab:confsep} shows the
result across all backbones and datasets, and Fig.~\ref{fig:confsep} plots it.

\begin{table}[t]
\centering
\caption{Exact Match of High- versus Low-Confidence Answers ($\tau = 0.6$)}
\label{tab:confsep}
\renewcommand{\arraystretch}{1.2}
\begin{tabular}{lcc}
\toprule
\textbf{Backbone / Dataset} & \textbf{High-conf. EM (n)} & \textbf{Low-conf. EM (n)} \\
\midrule
Qwen-1.5B / HotpotQA & 34.7\% (199) & 0.0\% (1) \\
Llama-3.2-3B / HotpotQA & 51.6\% (155) & 22.2\% (45) \\
Claude Haiku / HotpotQA & 36.5\% (181) & 0.0\% (19) \\
Claude Haiku / TriviaQA & 57.6\% (139) & 0.0\% (61) \\
\bottomrule
\end{tabular}
\end{table}

The separation is clear and it holds on every backbone. On the closed model with TriviaQA,
the high-confidence answers reach 57.6\% Exact Match while the low-confidence answers reach
zero, and the low-confidence group is large at sixty-one questions, so this is not a
small-sample artefact. On the closed model with HotpotQA the split is 36.5\% against zero,
and on the mid-sized Llama model it is 51.6\% against 22.2\%. The only weak case is the
small Qwen model, where almost every answer landed in the high-confidence group and only a
single question fell into the low-confidence group, which is too few to read anything
into. Setting that one aside, the finding is consistent: when the estimator reports high
confidence the answer is usually right, and when it reports low confidence the answer is
usually wrong. This is the result that makes AB-RAG useful for selective prediction,
because a system can answer when confident and abstain or retrieve more when not.

\subsection{The Cost-Accuracy Tradeoff}
\begin{figure*}[t]
\centering
\includegraphics[width=0.96\textwidth]{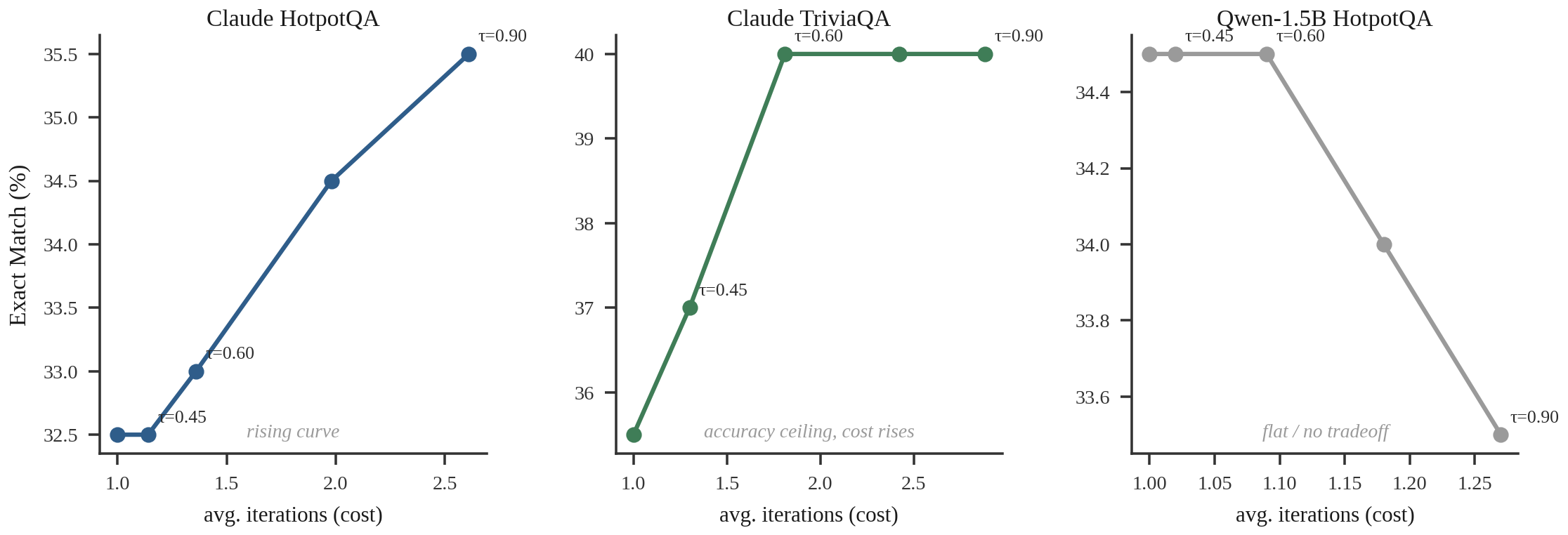}
\caption{Cost-accuracy tradeoff as the confidence threshold is swept. The closed model on
HotpotQA shows a rising curve, the closed model on TriviaQA holds accuracy while cost
rises, and the small Qwen model stays flat.}
\label{fig:tradeoff}
\end{figure*}

The adaptive loop exposes a single knob, the confidence threshold, that trades retrieval
cost against accuracy. Raising the threshold makes the system require more confidence
before it stops, so it retrieves more often and spends more, while lowering it makes the
system stop sooner and spend less. Fig.~\ref{fig:tradeoff} shows what happens to Exact
Match and to the average number of retrieval iterations as the threshold is swept, for
each backbone. Table~\ref{tab:sweep} gives the underlying numbers.

\begin{table}[t]
\centering
\caption{Confidence-Threshold Sweep: Avg. Iterations / Exact Match}
\label{tab:sweep}
\renewcommand{\arraystretch}{1.2}
\begin{tabular}{cccc}
\toprule
\textbf{$\tau$} & \textbf{Claude HotpotQA} & \textbf{Claude TriviaQA} & \textbf{Qwen HotpotQA} \\
\midrule
0.30 & 1.00 / 32.5 & 1.00 / 35.5 & 1.00 / 34.5 \\
0.45 & 1.14 / 32.5 & 1.30 / 37.0 & 1.02 / 34.5 \\
0.60 & 1.36 / 33.0 & 1.81 / 40.0 & 1.09 / 34.5 \\
0.75 & 1.98 / 34.5 & 2.42 / 40.0 & 1.18 / 34.0 \\
0.90 & 2.61 / 35.5 & 2.88 / 40.0 & 1.27 / 33.5 \\
\bottomrule
\end{tabular}
\end{table}

The shape of the curve depends on the backbone, and this is itself a finding. For the
closed model on HotpotQA, raising the threshold raises both the average number of
iterations and the Exact Match, producing the smooth cost-accuracy tradeoff the method
promises. For the closed model on TriviaQA, the Exact Match holds at its ceiling while the
cost rises with the threshold, which shows that the accuracy ceiling on that benchmark is
set by the backbone and the retrieval rather than by the policy, and that the policy's
value there is the quality of its confidence rather than a further lift in accuracy. For
the small Qwen model the curve is flat or slightly declining, confirming that a model must
be capable enough for the adaptive policy to give a real tradeoff. Taken together, the
curves show that AB-RAG gives a usable cost knob on capable backbones, and that the
benefit requires a backbone strong enough to make use of additional evidence.

\subsection{Signal Ablation and Diagnostics}
\begin{figure*}[t]
\centering
\includegraphics[width=0.92\textwidth]{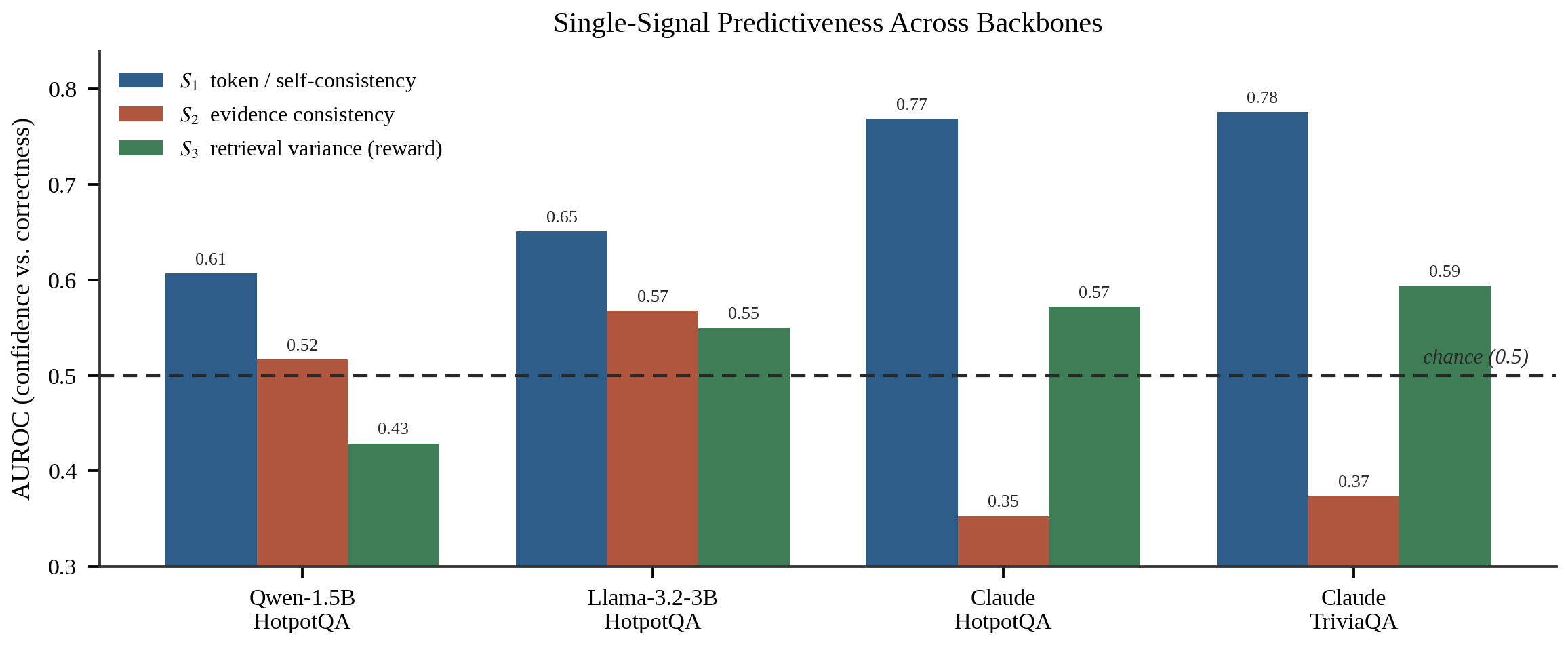}
\caption{Single-signal predictiveness across backbones. $S_1$, the model-certainty
signal, is strongly predictive everywhere; $S_2$, evidence consistency, is at or below
chance for these short answers; $S_3$, the corrected retrieval-variance reward, is weakly
predictive.}
\label{fig:auroc}
\end{figure*}

The confidence estimator combines three signals, and a fair question is whether all three
actually help. Rather than assume they do, each signal was treated as a hypothesis and
tested on its own. The measure used is the AUROC of that single signal against answer
correctness, which says how well the signal alone separates correct from incorrect
answers. Table~\ref{tab:auroc} reports the single-signal AUROC for all three signals on
every backbone and dataset, and Fig.~\ref{fig:auroc} plots it.

\begin{table}[t]
\centering
\caption{Single-Signal AUROC Against Correctness}
\label{tab:auroc}
\renewcommand{\arraystretch}{1.2}
\begin{tabular}{lccc}
\toprule
\textbf{Backbone / Dataset} & \textbf{$S_1$} & \textbf{$S_2$} & \textbf{$S_3$} \\
\midrule
Qwen-1.5B / HotpotQA & 0.607 & 0.517 & 0.429 \\
Llama-3.2-3B / HotpotQA & 0.651 & 0.568 & 0.550 \\
Claude Haiku / HotpotQA & \textbf{0.769} & 0.353 & 0.572 \\
Claude Haiku / TriviaQA & \textbf{0.776} & 0.374 & 0.594 \\
\bottomrule
\end{tabular}
\end{table}

\textit{Signal 1 is the workhorse.}
The model-certainty signal is strongly and consistently predictive. Its AUROC rises with
the capability of the backbone, from 0.607 on the small Qwen model to 0.769 and 0.776 on
the closed model. This is the clearest single result of the ablation: the better the
model, the more its own certainty tells us about whether it is right. This is why the
final weights give Signal~1 the largest share at $\alpha = 0.7$, and it is also why the
self-consistency proxy matters so much, because it is what carries this signal over to
closed models that do not expose token probabilities.

\textit{Signal 2 fails for short answers, and the reason is mechanistic.}
The evidence-consistency signal does not work for this task. Its AUROC sits around or below
one half on the closed model, at 0.353 and 0.374, which means it is no better than chance
and on these datasets slightly worse. This is a genuine negative result and it is reported
as one. The reason is mechanistic rather than accidental. The signal measures the cosine
similarity between the embedding of the answer and the embedding of the evidence passage,
but the answers in these datasets are very short, often a single name or a few words,
while the passages are long. Embedding a two-word answer and a hundred-word passage into
the same space and comparing them does not produce a meaningful grounding score, because
the two texts differ so much in length and specificity that the cosine similarity is
dominated by that mismatch rather than by whether the answer is supported. Several
alternative formulations were tried, and Table~\ref{tab:altsignal} shows that none of them
rescued the signal.

\begin{table}[t]
\centering
\caption{Alternative Formulations of $S_2$ and $S_3$ (Claude, HotpotQA, n=200)}
\label{tab:altsignal}
\renewcommand{\arraystretch}{1.2}
\begin{tabular}{lc}
\toprule
\textbf{Signal variant} & \textbf{AUROC} \\
\midrule
$S_2$: answer vs.\ mean-evidence & 0.347 \\
$S_2$: answer vs.\ best passage & 0.398 \\
$S_2$: answer vs.\ top-1 passage & 0.313 \\
$S_3$: $-$variance (penalty) & 0.427 \\
$S_3$: $+$variance (reward) & \textbf{0.573} \\
$S_3$: mean rerank score & 0.547 \\
\bottomrule
\end{tabular}
\end{table}

Because no formulation of the evidence-consistency signal reached a useful level, it is
kept in the estimator at a near-zero weight of $\beta = 0.05$ rather than dropped
entirely. Keeping it at a small weight does no harm and leaves the door open for tasks
with longer answers, where an answer-evidence similarity might carry more information.
Reporting this failure openly is more useful than quietly removing the signal, because it
tells anyone building a similar system that answer-evidence cosine similarity is the wrong
tool for short-answer grounding, and why.

\textit{Signal 3 had its sign backwards.}
The retrieval-variance signal produced the most instructive diagnostic of the project. In
the first implementation, the variance of the reranker scores was subtracted as a penalty,
on the intuition that spread-out scores meant the retriever was unsure. Measured on its
own, this penalty formulation gave an AUROC of 0.427, which is below chance, a strong hint
that the signal was pointing the wrong way. Inverting it, so that high variance is added
as a reward, raised the AUROC to 0.573, which is above chance. The corrected
interpretation is the right one: when the reranker assigns clearly different scores to
different passages, it has separated the relevant passages from the irrelevant ones with
confidence, and that clean separation is a sign of good retrieval. A flat, low-variance
score profile means the reranker could not tell the passages apart. This sign error was
found only because each signal was measured on its own rather than trusted, and correcting
it is a small but real example of measurement improving a design. The final estimator uses
the reward form with $\gamma = 0.25$.

\subsection{Qualitative Examples and Discussion}
Numbers alone can hide what a system is actually doing, so this section looks at concrete
outputs. Two patterns from the runs are worth showing directly: the verbosity effect that
depresses the closed model's Exact Match, and the adaptive loop doing its job on factoid
questions.

\textit{The verbosity effect.}
Table~\ref{tab:verbosity} shows three questions where the closed model gave an answer that
is correct in substance but was marked wrong by Exact Match because it included more than
the gold answer. In each case the model's answer contains the gold answer plus extra
context, such as a state name or a fuller phrasing. A human would mark all three correct.
This is the verbosity confound discussed earlier, and seeing the examples makes clear why
Exact Match understates the closed model and why the token-level F1 scores, which give
partial credit, are the fairer comparison across backbones.

\begin{table}[t]
\centering
\caption{Substantively Correct Answers Marked Wrong by Exact Match}
\label{tab:verbosity}
\renewcommand{\arraystretch}{1.2}
\begin{tabular}{llc}
\toprule
\textbf{Gold answer} & \textbf{Model answer} & \textbf{EM / F1} \\
\midrule
Wichita & Wichita, Kansas & 0 / high \\
French & the French language & 0 / high \\
1912 & in the year 1912 & 0 / high \\
\bottomrule
\end{tabular}
\end{table}

\textit{The adaptive loop on factoid questions.}
Table~\ref{tab:adaptive} shows the adaptive loop on TriviaQA questions, with the
confidence at the first round and the action it triggered. When the first answer is
already confident the loop stops after a single round and spends nothing extra. When the
first answer is not confident the loop retrieves more, and in several cases the second
round produces the correct answer that the first round missed. This is the intended
behaviour: cheap when the question is easy, willing to spend more when the question is
hard. It is also the behaviour behind the TriviaQA accuracy gain in
Table~\ref{tab:static}.

\begin{table}[t]
\centering
\caption{Adaptive Loop Behaviour on TriviaQA Questions}
\label{tab:adaptive}
\renewcommand{\arraystretch}{1.2}
\begin{tabular}{lll}
\toprule
\textbf{First-round confidence} & \textbf{Action} & \textbf{Outcome} \\
\midrule
0.83 (high) & stop at round 1 & correct, 1 round \\
0.30 $\rightarrow$ 0.77 & retrieve more & fixed in round 2 \\
0.41 $\rightarrow$ 0.79 & retrieve more & fixed in round 2 \\
0.22 (stays low) & stop at budget & abstain-worthy \\
\bottomrule
\end{tabular}
\end{table}

\textit{Discussion.}
Pulling the results together, a few broader points stand out. The first is that the value
of AB-RAG is not a single headline accuracy number but the reliability of its confidence.
On every backbone the confidence separated correct from incorrect answers, and that
property is what enables selective prediction, which is often more useful in practice than
a small average accuracy gain. The second is that the adaptive policy engages to different
degrees depending on the backbone, with the average number of iterations rising from 1.09
on the small model to 1.90 on the mid model and 1.81 on the closed model with TriviaQA.
The engagement is non-monotonic in raw model size, which suggests that how much a model
improves with more evidence depends on more than parameter count, including how the model
uses its context. The third is that several of the most useful findings here were negative
or corrective, the failed evidence signal and the inverted retrieval signal, and they were
only found because the signals were measured rather than assumed. The honest treatment of
these results is, in the view taken here, a feature of the work rather than a blemish.

\textit{Threats to validity.}
Several limits should be kept in mind when reading these results. The corpora are pooled
open-retrieval corpora of a few thousand passages rather than the full Wikipedia
collection, a deliberate choice forced by the single-laptop budget, so the absolute recall
numbers would differ at web scale even if the comparisons between methods are expected to
hold. The question counts are moderate, at five hundred for HotpotQA and two hundred for
TriviaQA, which is why bootstrap confidence intervals are reported alongside the point
estimates rather than left implicit. Exact Match penalises verbose but correct answers,
which is reported openly and softened by also reporting F1. Finally, the self-consistency
proxy uses three samples, and a larger number of samples might sharpen the Signal~1
estimate on closed models at additional cost. None of these limits undercuts the central
finding that confidence predicts correctness, but each marks a place where a larger study
could go further.

\section{Conclusion and Future Work}
\label{sec:conclusion}

\subsection{Conclusion}
This paper presented AB-RAG, a training-free and backbone-agnostic framework that makes
retrieval-augmented question answering adaptive and budgeted. Instead of retrieving a
fixed number of passages for every query, AB-RAG generates an answer, estimates its
confidence from three signals, and retrieves more evidence only when the answer is not yet
confident, subject to a fixed budget. The framework runs on both open-weight models, using
their real token probabilities, and closed APIs, using a self-consistency proxy in place
of those probabilities, which is what lets it apply to the many deployed systems built on
commercial models.

The central and most robust finding is that the confidence estimate reliably separates
correct from incorrect answers on every backbone tested, reaching a clean split of 57.6\%
against zero Exact Match between high- and low-confidence answers on a factoid dataset.
This makes the method directly useful for selective prediction, where a system answers
when confident and abstains or gathers more evidence when not. On capable backbones the
confidence threshold also acts as a single knob that trades retrieval cost against
accuracy, and on a paid API it doubles as a direct cost control because the budget bounds
the number of calls a query can cause. The study also reported its negative and corrective
findings in full: a small model that does not benefit from the adaptive policy, an
evidence-consistency signal that fails on short answers for a clear mechanistic reason,
and a retrieval-variance signal whose sign was found and corrected through measurement. The
entire study was completed on a single consumer laptop with a 4~GB GPU and only a few
dollars of API spend, which shows that careful, honest work on adaptive retrieval does not
require large resources.

\subsection{Future Work}
Several directions follow naturally from these results. The most direct is to scale the
retrieval corpus from the pooled open-retrieval setting used here to a full encyclopaedic
collection, to confirm that the confidence-correctness finding holds at web scale. A second
direction is to design a better grounding signal for short answers to replace the failed
evidence-consistency signal, for example a natural-language-inference check that asks
whether the evidence entails the answer, or a direct string-match against the passages,
both of which avoid the length-mismatch problem that defeated the cosine-similarity signal.
A third direction is to learn the signal weights rather than set them by hand, since the
ablation gives a clean supervised target in the form of single-signal AUROC. A fourth is
to broaden the evaluation to more backbones and to long-form generation tasks, where the
evidence-consistency signal might recover its value. A fifth is to study the latency of
the adaptive loop, since extra retrieval rounds cost wall-clock time as well as money, and
to characterise when the accuracy gain is worth the delay.

A further direction connects this work to the author's earlier study of adaptive,
decentralised decision-making in multi-agent systems~\cite{kamthan2025marl}. AB-RAG
currently makes its retrieve-or-stop decision with a single agent acting on its own
confidence. A natural extension is to treat retrieval, generation, and verification as
separate cooperating agents that share signals and decide jointly how much evidence to
gather, which would bring the budgeted-confidence idea developed here into the agentic,
multi-agent setting explored in that earlier work. Pursuing this would tie together the
two threads of adaptive decision-making under uncertainty: one in retrieval and one in
multi-agent control.

\section*{Acknowledgment}
The author thanks Dr.~Anamika Dhillon of the Department of Artificial Intelligence and
Machine Learning, Manipal University Jaipur, for her guidance and supervision throughout
this project.

\bibliographystyle{IEEEtran}
\bibliography{refs}

\end{document}